\documentclass[10pt,letterpaper]{article}

\usepackage[margin=1.9cm]{geometry}

\usepackage[none]{hyphenat}
\usepackage{hyperref}
\usepackage{graphicx}
\usepackage{caption}
\usepackage{subcaption}
\usepackage{listings}

\newenvironment{packed_itemize}{
\begin{itemize}
  \setlength{\itemsep}{1pt}
  \setlength{\parskip}{0pt}
  \setlength{\parsep}{0pt}
}{\end{itemize}}

\begin{document}

\title{Enabling more efficient and cost-effective AI/ML systems with Collective Mind, 
virtualized MLOps, MLPerf, Collective Knowledge Playground and reproducible optimization tournaments}

\author{Grigori Fursin\thanks{Email: grigori.fursin@ctuning.org ; gfursin@mlcommons.org}
\\
\href{https://cKnowledge.org}{cKnowledge.org} and \href{https://cTuning.org}{cTuning.org}
\\
\\
\href{https://github.com/mlcommons/ck}{github.com/mlcommons/ck}
\\
\href{https://github.com/mlcommons/ck/tree/master/cm/cmind}{github.com/mlcommons/ck/tree/master/cm/cmind}
\\
\href{https://access.cKnowledge.org/playground}{access.cKnowledge.org/playground}
}

\vskip 0.2in

\maketitle

\begin{abstract}

This white paper introduces my educational community initiative to learn 
how to run AI, ML and other emerging workloads in the most efficient 
and cost-effective way across diverse models, data sets, software and hardware.
This project leverages Collective Mind (CM), 
virtualized MLOps and DevOps (CM4MLOps), MLPerf benchmarks, 
and the Collective Knowledge playground (CK), which I have developed 
in collaboration with the community and MLCommons.

I created Collective Mind as a small and portable Python package with minimal dependencies, 
a unified CLI and Python API to help researchers and engineers automate repetitive, 
tedious, and time-consuming tasks, including those related to ResearchOps, MLOps, and DevOps.
I designed CM as a distributed framework, continuously
enhanced by the community through the CM4* repositories, which function
as the unified interface for organizing and managing various collections of automations and artifacts.
For example, \href{https://github.com/mlcommons/ck/tree/master/cm-mlops}{CM4MLOps repository}
includes \href{https://access.cknowledge.org/playground/?action=scripts}{many automations also known as CM scripts}
to streamline the process of building, running, benchmarking, and optimizing AI, 
ML, and other workflows across ever-evolving models, data, and systems.

I donated Collective Knowledge, along with prototypes of Collective Mind and CM4MLOps, 
to MLCommons to benefit everyone,  support the community and foster collaboration between academia and industry
to learn how to co-design more efficient and cost-effective AI systems while capturing and encoding 
knowledge within Collective Mind, protecting intellectual property, enabling portable skills, 
and accelerating the transition of the state-of-the-art research into production.

My ultimate goal is to collaborate with the community to complete my two-decade journey 
toward creating self-optimizing software and hardware that can automatically learn 
how to run any workload in the most efficient and cost-effective manner based on user requirements 
and constraints such as cost, latency, throughput, accuracy, power consumption, size, and other critical factors.

\end{abstract}

\vspace{0.3cm}

{\bf Keywords:}
{\it\small 
 open science, automation, reproducibility, reusability, modularity, virtualization, virtualized MLOps, virtual AI, 
 portability, collective mind, collective knowledge,  cknowledge, ctuning,
 best practices, ResearchOps, MLOps, DevOps, MLCommons, MLPerf, portable workflows, collaborative benchmarking, 
 cm4mlops, cm4abtf, cm4mlperf, cmx4mlops, cm scripts, optimization, software/hardware co-design, productivity, cost-effectiveness,
 open-source, artifact evaluation, competitions, hackathons, optimization challenges, education, portable skills
 self-learning, self-optimization
}

\vspace{0.3cm}

\section{Motivation}

Designing efficient and cost-effective computer systems to support emerging workloads is a highly complex, 
iterative, ad-hoc, and costly process~\cite{10.5555/3207796,Fur2009}.
Multiple OEMs, Tier 1 suppliers, system integrators, cloud providers and other organizations and individuals
must find a way to collaborate to connect, optimize and compare numerous and continuously changing ML models, 
datasets, software and hardware from different vendors and with different licenses while safeguarding 
the intellectual property (IP) of all parties involved.

For over two decades, I have engaged deeply with these challenges while developing a machine-learning-based compiler 
featuring collective tuning and federated learning, spearheading the artifact evaluation process 
with a standardized artifact appendix and reproducibility checklists for ACM and IEEE conferences, 
reproducing results from numerous academic and industry research projects, 
and assisting companies and the broader community in running MLPerf benchmarks and submitting 
highly efficient and cost-effective software and hardware configurations optimized for popular AI and ML tasks
~\cite{acm_techtalk_fursin_reproducibility_2022,Fur2009,29db2248aba45e59:a31e374796869125,cfkz2016}. 

In my work with academia and industry, I observed that preparing,
documenting, reproducing, and debugging complex experiments often requires
weeks or even months of tedious, ad-hoc, and repetitive interactions among
teams from different organizations. 
These experiments typically involve
empirical measurements of runtime characteristics such as latency,
throughput, and energy consumption, which are highly variable and
challenging to manage.
This complex process involves writing some ad-hoc README files, Jupyter notebooks and spreadsheets, along with some ad-hoc containers
and instructions how to run experiments. These artifacts are then shared among teams, who often spend significant amounts of time 
parsing and updating them to reproduce empirical results across the ever-evolving landscape of software and hardware~\cite{mad_2024}.

This hands-on and often tedious experience inspired me to explore
virtualizing the process of building, running, benchmarking, and
optimizing AI, ML, and other complex workloads. The goal was to abstract
these workflows from the ever-changing models, datasets, software, and
hardware - much like how hardware virtualization and containers enable the
seamless operation of different operating systems and applications across
diverse hardware configurations.
Another objective I pursued was to create a unified language for
automation and reproducibility, enabling seamless access to all research
projects, with a particular focus on co-designing and optimizing AI and ML
systems.

As a result of this research, I developed a prototype automation framework 
that I called Collective Mind (CM)~\cite{acm_rep_23_cm_keynote,github_cmind} -
a lightweight Python-based tool to assist researchers and engineers in automating, sharing, 
and reusing repetitive, tedious, and time-consuming tasks 
in a standardized and portable manner.
I also designed CM to be a distributed framework, continuously
enhanced by the community through the CM4* Git repositories.

To validate my Collective Mind concept, I developed a prototype 
\href{https://github.com/mlcommons/ck/tree/master/cm-mlops}{CM4MLOps repository}
featuring two key components: a CM script and a CM cache. These provide
a common, technology-agnostic approach to virtualizing ResearchOps, MLOps,
and DevOps, enabling their integration into portable workflows. These
workflows facilitate the building, execution, benchmarking, and
optimization of emerging workloads across a wide range of continuously
evolving models, datasets, software, and hardware from various vendors
in a more unified, automated, and reproducible manner~\cite{cm4mlops,cm_scripts}.

CM enables teams across different companies to collaborate 
using a common, human-readable and user-friendly automation language.
It also supports the co-design, optimization, and comparison of software 
and hardware while allowing the plug\&play integration of public or private components.
Additionally, it automates the generation of container snapshots and READMEs 
with standardized CM commands, ensuring license compliance and safeguarding intellectual property.

However, as CM users faced challenges in locating, managing, and testing
the increasing number of automation recipes and workflows dispersed across
various CM repositories, I developed the \href{https://access.cKnowledge.org/playground}{Collective Knowledge Playground also known as CK playground}.
This open platform is designed to index all automation recipes similar to PYPI~\cite{cm_scripts},
assist users in preparing CM commands to run MLPerf inference benchmarks~\cite{cknowledge_howtorun},
aggregate, process, visualize, and compare benchmarking results for
AI and ML systems~\cite{cknowledge_experiments}, and organize open, reproducible optimization challenges
and tournaments. These initiatives aim to help academia and industry
collaboratively enhance the efficiency and cost-effectiveness of AI
systems~\cite{cm_challenges}.

I also collaborated with Arjun Suresh to develop a modular and technology-agnostic
CM workflow for running MLPerf inference benchmarks~\cite{cm4mlperf} and
organized the first open challenge to benchmark commodity software and
hardware for AI using MLPerf and CM.
This challenge showcased the power of Collective Mind by enabling the
first large-scale community submission of 12,217 inference benchmark v3.1
results, representing over 90\% of all power measurements, out of a total
of 13,351 results contributed by 20+ AI companies. These submissions
spanned diverse servers, embedded devices, and cloud instances featuring
processors and accelerators from vendors such as Nvidia, Qualcomm, Intel,
AMD, Amazon, Broadcom, and
others~\cite{democratize_mlperf_v3_1,new_cm_workflow_for_mlperf_4_0}, made
possible through the efforts of our dedicated
volunteers~\cite{cm_volunteers}.

I hope that my Collective Mind technology, in combination with the
\href{https://access.cKnowledge.org/playground}{Collective Knowledge
Playground} and MLPerf, will help democratize AI research, development,
and education, reduce the costs associated with building, running, and
optimizing AI, and make AI accessible to everyone.

I continue collaborating with
MLCommons~\cite{mlcommons_launches,cm_avcc_mlperf,cm4abtf,6e86adf711484875971b8586a02af1b7},
ACM/IEEE conferences~\cite{ae_micro23_cm_report,cm4research}, the Student
Cluster Competition at the SuperComputing conference~\cite{cm4scc23}, and
the wider community to support students, researchers, and engineers.
My aim is to enable them to encode their knowledge and expertise into
CM automation recipes and artifacts, empowering them to focus
on innovation and optimization while seamlessly adapting to the rapid
evolution of models, software, and hardware.

One of my current objectives is to automate, virtualize, and streamline
the software/hardware co-design of complex AI and ML systems, transforming
them into a commodity that can be automatically assembled from the most
suitable open-source and proprietary components from diverse vendors. 
Such process should be guided by user demands, requirements, and constraints such
as cost, latency, throughput, accuracy, energy efficiency, size, and other
key factors.

\section*{Collective Mind (CM): a toolset to create and share portable automations}

\subsection*{Collective Mind architecture}

\begin{figure*}[]
  \centering
  \includegraphics[width=1.0\textwidth]{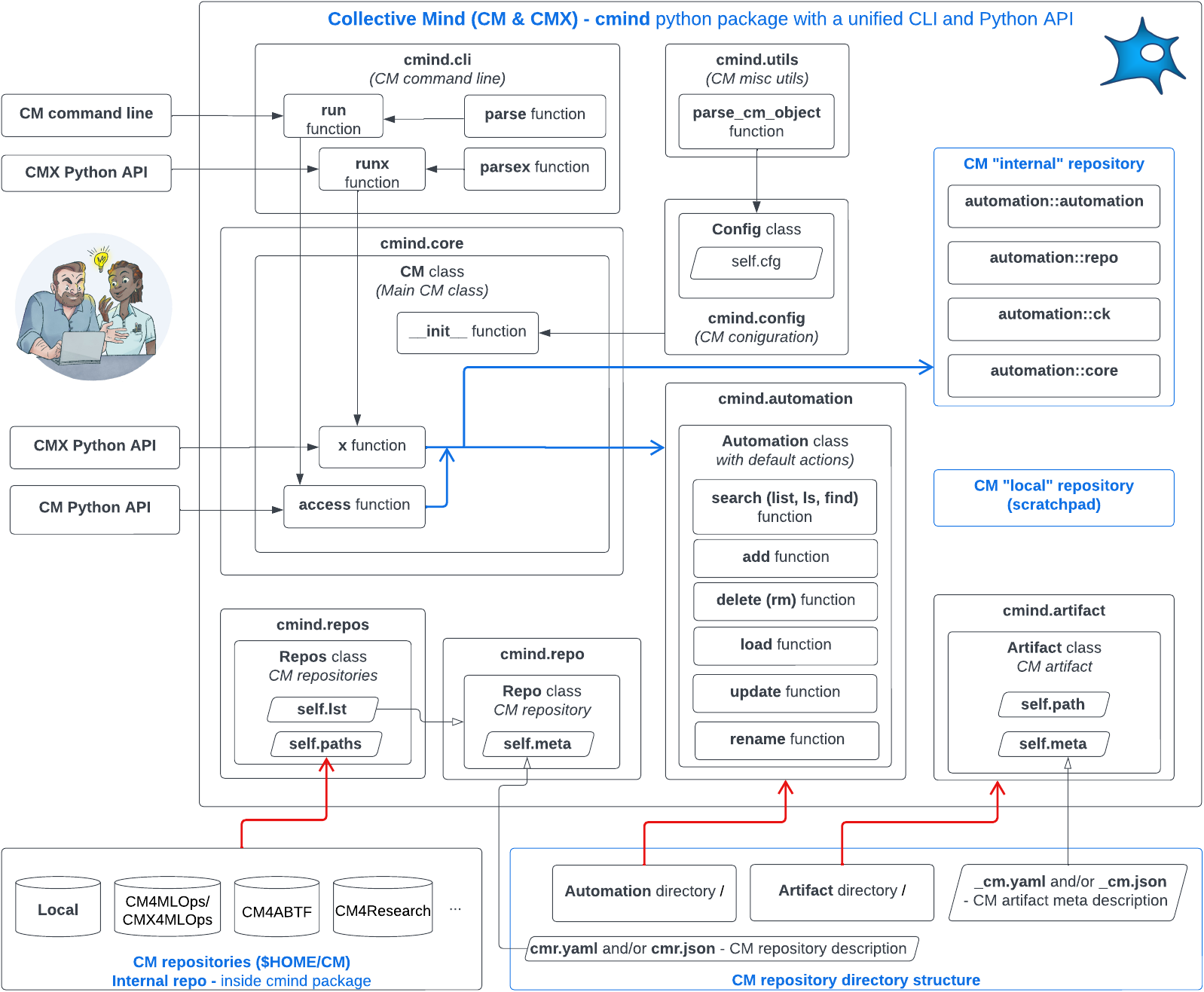}
  \caption{The decentralized architecture of Collective Mind.}
  \label{fig:cm-architecture}
\end{figure*}

I designed CM as a decentralized Python-based automation framework, extendable through public and private Git repositories.

The diagram Figure~\ref{fig:cm-architecture} illustrates the architecture of \textbf{Collective Mind (CM)}. 
It provides an overview of its modular structure and how the components interact with each other. 
Below is a breakdown of the key elements:

\textbf{CM Command Line (CLI) and Python API}
\begin{itemize}
    \item \textbf{User Interfaces:}
    \begin{itemize}
        \item The CM framework can be accessed either via a command-line interface (CLI) or programmatically through Python.
        \item CLI users invoke commands using functions such as \texttt{run} and \texttt{parse} within the \texttt{cmind.cli} module.
        \item Python users interact with the CM framework via the \texttt{access} function within the main \texttt{CM} class in \texttt{cmind.core}.
    \end{itemize}
\end{itemize}

\textbf{Core CM Framework}
\begin{itemize}
    \item \textbf{\texttt{cmind.core} Module:}
    \begin{itemize}
        \item At the heart of the framework is the \texttt{CM} class, which is initialized using the \texttt{\_\_init\_\_} function.
        \item This class coordinates access to other components such as repositories, automation, and artifacts.
        \item The \texttt{access} function provides a unified entry point for accessing various CM functionalities.
    \end{itemize}
    \item \textbf{\texttt{cmind.config} and \texttt{cmind.utils}:}
    \begin{itemize}
        \item The \texttt{Config} class (\texttt{self.cfg}) manages configuration settings for the framework.
        \item Utility functions, like \texttt{parse\_cm\_object}, handle miscellaneous functions commonly found in DevOps and MLOps projects~\cite{cm_api}.
    \end{itemize}
\end{itemize}

\textbf{CM Repositories and Automation}
\begin{itemize}
    \item \textbf{Repositories (\texttt{cmind.repos} and \texttt{cmind.repo}):}
    \begin{itemize}
        \item CM repositories are directories where CM artifacts and automation scripts are stored.
        \item CM always have two repositories after installation:
        \begin{itemize}
            \item \textbf{Local:} CM scratchpad repository.
            \item \textbf{Internal:} Built-in repository, provided within the \texttt{cmind} package.
        \end{itemize}
        \item The \texttt{Repos} class (\texttt{self.lst} and \texttt{self.paths}) manages a list of repositories and their paths.
        \item The \texttt{Repo} class (\texttt{self.meta}) represents metadata for individual repositories.
    \end{itemize}
    \item \textbf{Automation (\texttt{cmind.automation}):}
    \begin{itemize}
        \item This module manages automated processes and actions.
        \item The \texttt{Automation} class includes methods for operations like:
        \begin{itemize}
            \item \texttt{search} (list or find objects).
            \item \texttt{add}, \texttt{delete}, \texttt{update}, and \texttt{rename} objects.
            \item \texttt{load} artifacts or automation components from repositories.
        \end{itemize}
    \end{itemize}
\end{itemize}

\textbf{CM artifacts}
\begin{itemize}
    \item \textbf{Artifact Management (\texttt{cmind.artifact}):}
    \begin{itemize}
        \item CM artifacts represent reusable components stored within repositories.
        \item Artifacts are described using metadata files (\texttt{\_cm.yaml} or \texttt{\_cm.json}).
        \item The \texttt{Artifact} class manages artifact paths (\texttt{self.path}) and metadata (\texttt{self.meta}).
    \end{itemize}
\end{itemize}

\textbf{CM Repository Structure}
\begin{itemize}
    \item \textbf{Key Directory Types:}
    \begin{itemize}
        \item \textbf{Automation Directory:} Contains automation scripts or workflows.
        \item \textbf{Artifact Directory:} Contains CM artifacts and associated metadata files.
    \end{itemize}
\end{itemize}

\textbf{CM Repositories and Metadata Files}
\begin{itemize}
    \item The repositories follow a structured organization:
    \begin{itemize}
        \item Metadata files like \texttt{cmr.yaml} or \texttt{cmr.json} describe the repository contents.
        \item Artifacts and automation objects are defined by their own \texttt{\_cm.yaml} or \texttt{\_cm.json} metadata.
    \end{itemize}
\end{itemize}

\textbf{CM Internal and Local Repositories}
\begin{itemize}
    \item \textbf{Internal Repository:}
    \begin{itemize}
        \item Built into the framework, it contains the following core automation and configuration components:
        \begin{itemize}
            \item \texttt{automation::automation}
            \item \texttt{automation::repo}
            \item \texttt{automation::ck}
            \item \texttt{automation::core}
        \end{itemize}
    \end{itemize}
    \item \textbf{Local Repository:}
    \begin{itemize}
        \item A temporary workspace for development or customization.
    \end{itemize}
\end{itemize}

\textbf{Interaction Overview}
\begin{itemize}
    \item The framework is modular, allowing components to interact efficiently:
    \begin{itemize}
        \item The \texttt{access} function in \texttt{cmind.core} acts as a bridge between Python code, the CM CLI, and other modules.
        \item CM repositories organize and manage reusable automation workflows and artifacts.
        \item Users can easily search, modify, and retrieve artifacts and workflows using automation functions.
    \end{itemize}
\end{itemize}

This architecture ensures flexibility, scalability, and modularity for portable automation and distributed artifact management.

The Python API of the CM framework is available at~\href{https://cknowledge.org/docs/cm}{cKnowledge.org/docs/cm}.

\subsection*{Collective Mind Command Line}

A primary objective of the CM language is to offer a standardized,
unified, and human-readable CLI for accessing all software repositories
shared in the CM format.

\subsection{Format}

The goal is to standardize all shared READMEs, containers, and Jupyter
notebooks using CM commands, simplifying the process for the community
to run software projects and reuse individual automations in an
environment of constantly evolving software, hardware, and data.

Below is the format of a unified CM command line for executing any reusable CM automation 
from any software project on Linux, Windows, or macOS:

\begin{verbatim}
cm {action} {automation alias | UID | alias,UID} 
  ({artifact name(s) | sub-action | argument}) 
  (--flag1=value1) (--flag2.key2=value2) (--flag3,=value3,value4) (--flag4)
  (@input.json | @input.yaml)
  (-- extra CMD)
\end{verbatim}

Initially, CM will
\href{https://github.com/mlcommons/ck/blob/master/cm/cmind/cli.py#L54}{parse
the CM CLI} into a unified CM input dictionary:

\begin{verbatim}
{
  "action": "automation action",
  "automation": "automation alias | UID | alias,UID",
  "artifact": {above artifact name or sub-action},
  "flag1": "value1",
  "flag2": {"key2": "value2"},
  "flag3": ["value3", "value4"],
  "flag4": true,
  ...
  "unparsed_cmd": [
    list of strings in extra CMD
  ]
}
\end{verbatim}

When a user specifies one or more input files using the \texttt{@} prefix,
these files will be loaded and merged with the CM input in the same order
they appear in the command line.

CM will then invoke a
\href{https://github.com/mlcommons/ck/blob/master/cm/cmind/core.py#L138}{unified
CM Python "access" function} using this input dictionary to execute the
specified automation action.

This process is analogous to using the \href{cm-python-interface.md}{CM
Python API}, with the distinction that CM operates in interactive mode.
You can include the flag \texttt{--out=json} to display the output
dictionary at the conclusion of an automation action executed through the
CLI.

You can test the CM interface by using the following automation action, which simply prints the unified CM input dictionary:

\begin{verbatim}
cm print-input automation artifact1 artifact2 --flag1=value1 --flag2 -- something
\end{verbatim}

\subsection{Examples}

\begin{verbatim}
cm 
cm help
cm {common automation action}
\end{verbatim}

\begin{verbatim}
cm {common automation action} --help
\end{verbatim}

\begin{verbatim}
cm {automation action} {automation}
\end{verbatim}

\begin{verbatim}
cm {automation action} {automation} --help
\end{verbatim}

\begin{verbatim}
cm {automation action} {automation} {artifact}
\end{verbatim}

\begin{verbatim}
cm {automation action} {automation} {artifact} {artifact2} {artifact3} ...
\end{verbatim}

\begin{verbatim}
cm {automation action} {automation} {artifact} --test --meta.a=b @input.json @input.yaml
\end{verbatim}

Command line arguments are transformed into a unified CM dictionary
through
\href{https://github.com/mlcommons/ck/blob/master/cm/cmind/cli.py#L48}{this
function}.

Flags are mapped to dictionary keys, and their arguments are converted
to string values. If a flag does not have an argument, CM assigns it
a Boolean value of \texttt{true}. Flags containing .'' are interpreted
as dictionaries with subkeys separated by .''. If a flag ends with ``,''
its arguments are treated as a list of values separated by commas.

The CM dictionary is subsequently passed to the
\href{https://github.com/mlcommons/ck/blob/master/cm/cmind/core.py#L134}{unified
CM "access" function}, functioning similarly to microservices and REST APIs.

\subsection{Understanding CM Naming Conventions}

The \texttt{CM \{artifact\}} has the following formats:
\begin{itemize}
  \item \textbf{``artifact alias''} (string): May change in the future.
  \item \textbf{``artifact UID''} (16 lowercase hex digits): Remains constant since the artifact's creation.
  \item \textbf{``alias,UID''}: In this context, the UID is utilized to locate the artifact or automation, 
      whereas the alias acts as a user-friendly identifier.

\end{itemize}

Artifacts within specific CM repositories can be referenced using the following formats:

\begin{itemize}
  \item \texttt{repo alias:artifact}
  \item \texttt{repo UID:artifact}
  \item \texttt{repo alias,repo UID:artifact}
\end{itemize}

Automation uses the same format as artifacts and is stored as a CM
artifact within the \texttt{automation} directory. Similarly,
CM repositories follow the same naming format: \texttt{alias | UID | alias,UID}.

\subsection{Using CM CLI within a CM Repository}

When working within a CM repository, you can use \texttt{.} to enable
CM to automatically detect the repository, automation, and artifact in the
current directory, eliminating the need to explicitly specify their names
in the command line.

For instance, to add a new artifact in the current repository, use:
\begin{verbatim}
cm add {some automation} .:{some artifact}
\end{verbatim}

Or, if inside a directory with CM automations:
\begin{verbatim}
cm add . {some artifact}
\end{verbatim}

\section*{CM4MLOPs: portable and technology-agnostic automation recipes for MLOps, DevOps and ResearchOps}

\begin{figure*}[]
  \centering
  \includegraphics[width=1.0\textwidth]{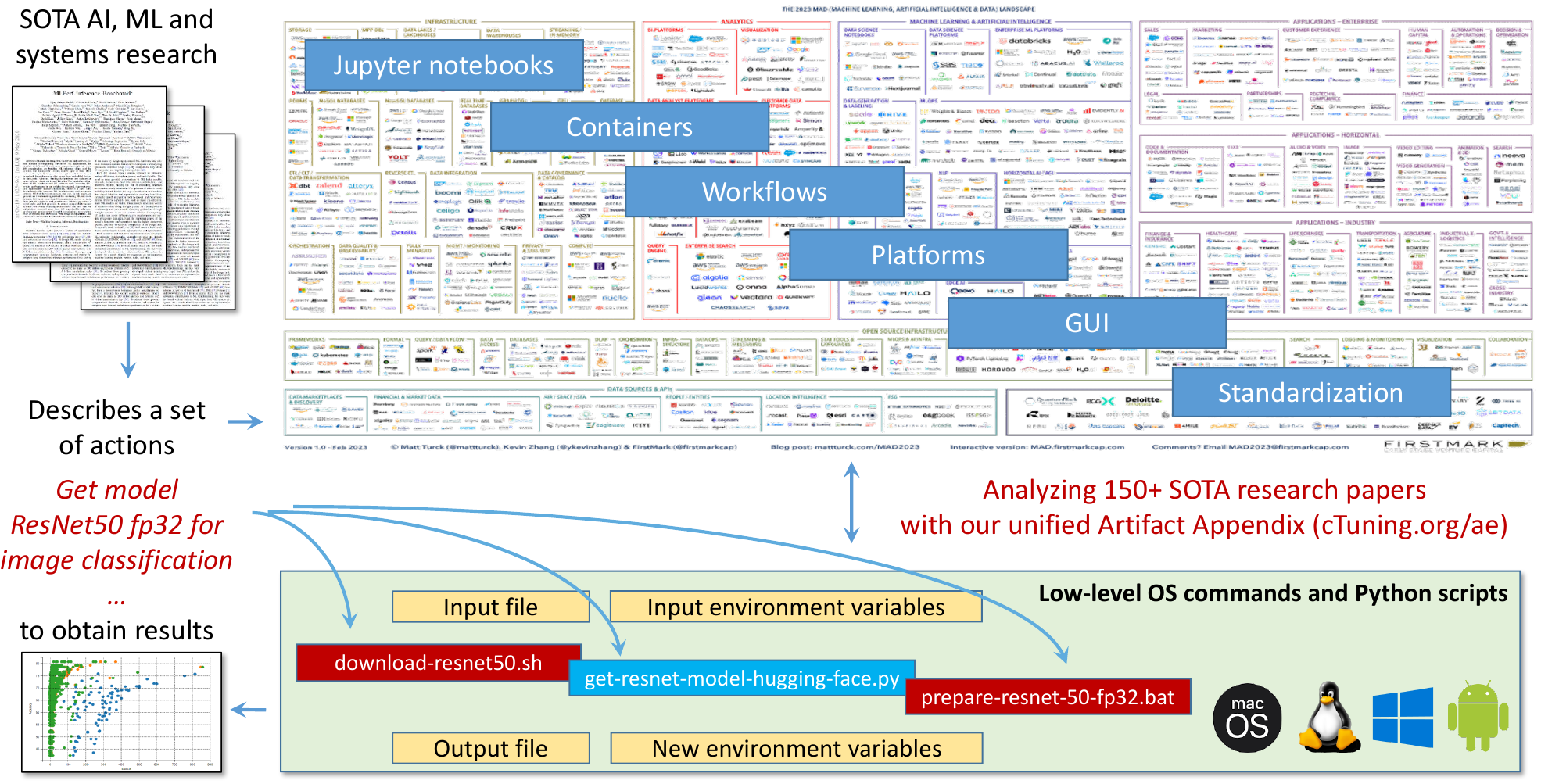}
  \caption{Many research projects implement the same "research actions"
  using diverse ad-hoc OS commands and Python scripts, sharing them along
  with loosely organized Git repositories, containers, and Jupyter
  notebooks to enable the community to reuse their projects and reproduce
  experiments~\cite{doi:10.1098/rsta.2020.0211,acm_techtalk_fursin_reproducibility_2022}.
  \textit{This figure includes the MAD landscape (Machine Learning, AI,
  and Data) sourced from
  \href{https://mad.firstmark.com}{mad.firstmark.com}}.
  }
  \label{fig:ad-hoc-scripts}
\end{figure*}

While reproducing over 150 state-of-the-art research papers presented
at AI, ML, and systems conferences~\cite{acm_techtalk_fursin_reproducibility_2022}, 
I observed that most researchers and
engineers develop ad-hoc automation solutions from scratch to perform
identical tasks across various tools, models, datasets, software, and
hardware. These tasks often include setting up environment variables,
updating file paths, managing build dependencies and applications,
creating containers, and re-running experiments. Typically, these are
implemented using ad-hoc, low-level, and OS-specific commands and scripts,
as illustrated in Figure~\ref{fig:ad-hoc-scripts}.

\begin{figure*}[]
  \centering
  \includegraphics[width=0.5\textwidth]{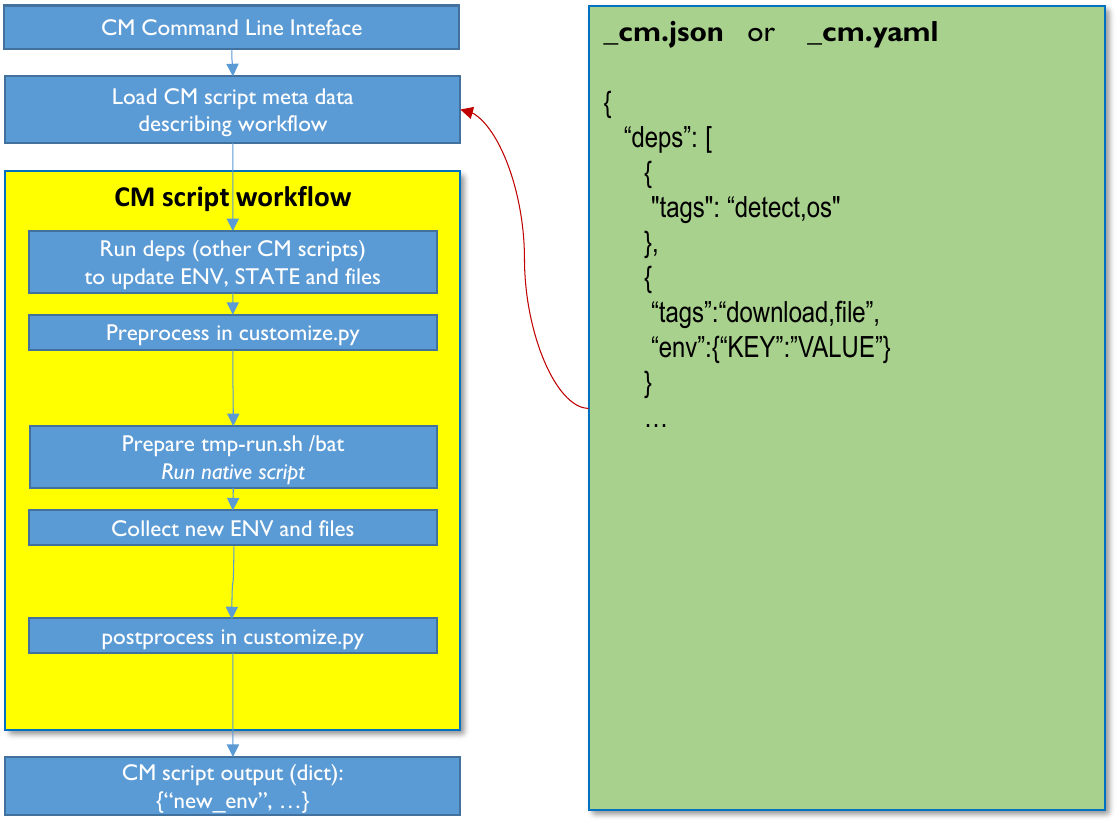}
  \caption{The basic workflow of a CM script}
  \label{fig:cm-script-flow}
\end{figure*}

\begin{figure*}[]
  \centering
  \includegraphics[width=1.0\textwidth]{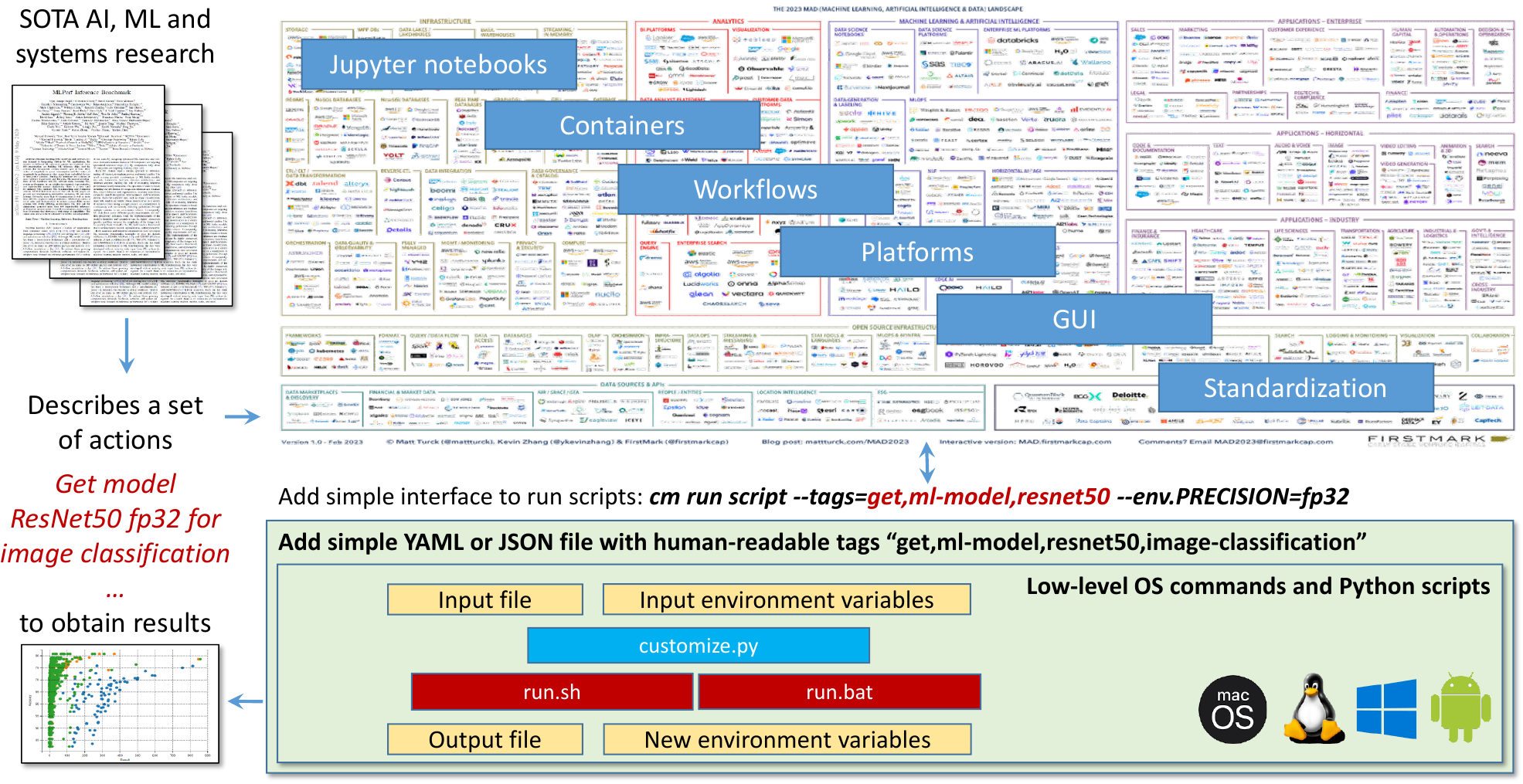}
  \caption{The Collective Mind framework enables the creation of portable, 
  technology-agnostic automation recipes (CM scripts). These scripts can be seamlessly reused 
  across diverse research projects, dynamically adapting to varying operating systems, models, 
  datasets, software, and hardware.
  }
  \label{fig:cm-scripts}
\end{figure*}

This challenging experience inspired me to develop a technology-agnostic CM automation called the CM script. 
CM scripts wrap native scripts and OS commands, enabling them to run in a unified manner across 
any operating system, software, or hardware~\ref{fig:cm-script-flow}. They provide a platform-independent command line 
or Python API with a human-readable interface—an intermediate automation language—featuring 
intuitive commands like \texttt{get ml-model resnet50} and a customizable Python module~\ref{fig:cm-scripts}.

This tedious experience inspired me to create a technology-agnostic automation called CM script
that wraps native scripts and OS commands and run them in a unified way on any Operating system with any software and hardware
using a platform-independent command line or Python API with a human-readable interface—an intermediate automation 
language—featuring commands like \texttt{get ml-model resnet50} and a customizable Python module to resolve dependencies
on other CM scripts, set up environment variables, command lines and adapt execution for a given platform.

Additionally, CM scripts can cache artifacts generated during execution
using CM Cache, a complementary automation tool from the CM4MLOps
repository.

To some extent, CM builds upon the concepts of the popular CMake build
system~\cite{cmake}, but extends its functionality with reusable,
technology-agnostic automation recipes and portable workflows. These
workflows are written in plain Python or native OS scripts, accessible
through a human-readable interface with simple tags, and shareable in both
public and private Git repositories or archives in a decentralized manner.
However, unlike CMake, CM automation recipes can manage not only code
but also models, datasets, and experiments. They can automatically
download or integrate missing artifacts (such as models, datasets, and
tools), preprocess them for a specific platform and environment, resolve
and build missing dependencies, install them, execute the final code, and
process the output—all in a unified, automated, and platform-independent
manner.
Furthermore, CM provides a simple command-line interface with a Python
API, enabling users to share, run, and reuse automation recipes stored
in GitHub repositories, TAR/ZIP files, containers, and PyPI packages:

\begin{lstlisting}[language=bash, basicstyle=\footnotesize]

$ pip install cmind

$ cm pull repo mlcommons@cm4mlops

List/find scripts by UID, alias and tags:

$ cm find script
$ cm ls script
$ cm find script 5b4e0237da074764
$ cm find script *-ml-model-*
$ cm find script --tags=resnet50

Load meta description of a given script:

$ cm load script get-ml-model-resnet50 --json

Add and run dummy script:

$ cm add script my-new-cool-script --tags=my,new,cool-script
$ cm run script --tags=my,new,cool-script --env.KEY=VALUE --json

Delete script:

$ cm rm script --tags=my,new,cool-script
$ cm delete script --tags=my,new,cool-script

...

\end{lstlisting}

I also introduced a new functionality in CM4MLOps to cache the output
of CM scripts using the CM Cache automation, enabling them to be managed
as a local file-based repository.

\begin{lstlisting}[language=bash, basicstyle=\footnotesize]

cm run script "get ml-model resnet50 _fp32 _onnx" --json

cm list cache
cm find cache --tags=ml-model,resnet50,_fp32
cm rm cache --tags=ml-model

cm rm cache -f
cm run script "get ml-model resnet50 _fp32 _onnx" --verbose

cm show repo

...

\end{lstlisting}

You can find more information about the CM command line, Python API, and architecture in the CM documentation~\cite{cm_docs}.

\section*{CM scripts}

As a starting point, I developed a prototype collection of CM scripts to address the most commonly used MLOps and DevOps 
tasks I encountered while developing machine-learning based compilers,
crowd-benchmarking, crowd-tuning, and federated learning, 
automating system co-design, reproducing research papers, 
and collaborating with MLCommons and OctoAI—hence the name CM4MLOps for this repository.
While prototyping CM scripts, I gathered the following requirements from researchers and engineers:

\begin{itemize} 

\item Must function out of the box with default options, eliminating the need to manually edit paths, environment variables, or configuration files;
\item Must be non-intrusive, easy to debug, and capable of reusing existing user scripts and automation tools (such as CMake, Make, Bazel, MLFlow, PIP, Poetry, Docker) rather than replacing them;
\item Must provide a simple, human-friendly command-line interface, accompanied by a Python API, and require minimal dependencies;
\item Must have minimal or zero learning curve by leveraging plain Python, native scripts, environment variables, and straightforward JSON/YAML descriptions, avoiding the creation of new workflow languages;
\item Must provide a consistent interface to run all automations natively, in the cloud, or within containers.

\end{itemize}

Here is an example of a CM script designed to prepare and run a simple image classification task with ONNX run-time, 
ResNet50 and ImageNet labels.
This automation recipe executes in a unified way across Linux, Windows, macOS, RHEL, Debian, Amazon Linux, and other platforms:

\begin{lstlisting}[language=bash, basicstyle=\footnotesize]

cm run script "detect os" --out=json
cm run script "get python" --version_min=3.9.1
cm run script "get ml-model resnet50 _onnx _fp32"
cm run script "get original imagenet dataset _2012-500"
cm run script "get generic-python-lib _onnxruntime"

cm run script "download file _wget" --url=https://cKnowledge.org/ai/data/computer_mouse.jpg \
     --verify=no --env.CM_DOWNLOAD_CHECKSUM=45ae5c940233892c2f860efdf0b66e7e

cm run script "python app image-classification onnx _cpu" --input=computer_mouse.jpg

\end{lstlisting}

Note that the final CM script is implemented as a straightforward and
portable pipeline (workflow) that integrates all the preceding CM scripts.
This is achieved using a simple YAML description with defined dependencies
on other CM scripts using simple tags:

\begin{lstlisting}[language=bash, basicstyle=\footnotesize]

deps:
- tags: detect,os
- tags: get,sys-utils-cm
- names:
  - python
  - python3
  tags: get,python3

- tags: get,cuda
  names:
  - cuda
  enable_if_env:
    USE_CUDA:
    - yes
- tags: get,cudnn
  names:
  - cudnn
  enable_if_env:
    USE_CUDA:
    - yes

- tags: get,dataset,imagenet,image-classification,original
- tags: get,dataset-aux,imagenet-aux,image-classification
- tags: get,ml-model,resnet50,_onnx,image-classification
  names:
  - ml-model

- tags: get,generic-python-lib,_package.Pillow
- tags: get,generic-python-lib,_package.numpy
- tags: get,generic-python-lib,_package.opencv-python


- tags: get,generic-python-lib,_onnxruntime
  names:
  - onnxruntime
  skip_if_env:
    USE_CUDA:
    - yes
- tags: get,generic-python-lib,_onnxruntime_gpu
  names:
  - onnxruntime
  enable_if_env:
    USE_CUDA:
    - yes

\end{lstlisting}
                           
When this pipeline is run again, all models, datasets, and packages will be cached and automatically integrated 
into the workflow using CM cache automation, significantly speeding up execution.
Furthermore, all cached artifacts will be available for reuse by any other
CM script or pipeline, similar to how Docker reuses layers between
containers.

The modular structure of the CM framework enables other useful features:

\begin{packed_itemize} 
\item running CM scripts in an automatically generated container by simply
replacing \texttt{cm run script} with \texttt{cm docker script}. This
reduces the burden on users by eliminating the need to create and manage
numerous ad-hoc containers.
\item running, recording and replaying experiments (such as benchmarking
and optimization) in a more deterministic 
and reproducible way using \texttt{cm run experiment} and \texttt{cm replay experiment}.
\end{packed_itemize}

After donating CK and CM to MLCommons, I also established the Automation
and Reproducibility Task Force~\cite{mlcommons_tf_automation_reproducibility} 
and led community efforts to extend
existing CM scripts or develop new ones to cover the following
categories~\cite{cm_scripts,cm_getting_started_guide}:

\begin{packed_itemize} 
\item AI/ML datasets
\item AI/ML frameworks
\item AI/ML models
\item AI/ML optimization
\item Cloud automation
\item CM automation
\item CM Interface
\item Collective benchmarking
\item Compiler automation
\item CUDA automation
\item Dashboard automation
\item Detection or installation of tools and artifacts
\item DevOps automation
\item Docker automation
\item MLPerf benchmark support
\item Modular AI/ML application pipeline
\item Modular application pipeline
\item Modular MLPerf benchmarks
\item Modular MLPerf inference benchmark pipeline
\item Modular MLPerf inference benchmark pipeline for ABTF model
\item Platform information
\item Python automation
\item Remote automation
\item Reproduce MLPerf benchmarks
\item Reproducibility and artifact evaluation
\item TinyML automation
\item Unsorted
...
\end{packed_itemize}

\section*{CM4MLPerf: modularizing and automating MLPerf benchmarks}

As a founding member of MLCommons~\cite{mlcommons_launches}, one of the
reasons I donated CK and CM to MLCommons is to leverage this open-source
technology to modularize MLPerf benchmarks. This will simplify running and
reproducing them across diverse hardware and software from vendors like Qualcomm,
Intel, Nvidia, AMD, Neural Magic and others, while also assisting everyone
in automating their submissions.

Within the MLCommons Task Force on Automation and Reproducibility, and in
collaboration with Arjun Suresh, I led the development of a collection
of portable CM scripts to modularize and run MLPerf inference benchmarks,
as well as measure power consumption across diverse systems in a unified
manner~\cite{cm4mlperf,cm4mlperf_mlcommons_docs}.

I also organized the first open challenge to benchmark commodity software and hardware 
using Collective Mind and MLPerf inference benchmarks~\cite{cm_challenges}.
This challenge enabled the first mass-scale community submission of 12,217 inference benchmark v3.1 results 
out of total 13,351 results from 20+ AI companies
(including 90\% of all power results).
These submissions served as a compelling proof-of-concept for my Collective Mind concepts, 
spanning diverse servers, embedded devices, and cloud instances with
processors and accelerators from Nvidia, Qualcomm, Intel, AMD, Amazon,
Broadcom, and other vendors~\cite{democratize_mlperf_v3_1,new_cm_workflow_for_mlperf_4_0}, 
all made possible by the contributions of our dedicated
volunteers~\cite{cm_volunteers}.

At the same time, I was honored to help bootstrap the use of MLPerf with
CM automations at the Student Cluster Competition in collaboration, 
starting from 2022~\cite{sc22-scc-mlperf}.

In SCC 2024, we used the Stable Diffusion XL model, which can be run on a system
with Nvidia GPUs via CM with a single command
line~\cite{ctuning-mlperf-inference_v4.1-submission}:

\begin{lstlisting}[language=bash, basicstyle=\footnotesize]

pip install -U cmind

cm pull repo mlcommons@cm4mlops

cm run script \
	--tags=run-mlperf,inference,_r4.1 \
	--model=sdxl \
	--implementation=reference \
	--framework=pytorch \
	--category=datacenter \
	--scenario=Offline \
	--execution_mode=valid \
	--device=cuda \
	--quiet

\end{lstlisting}

Furthermore, we collaborated with the community to prototype and test
a modular C++ implementation of the MLPerf benchmark~\cite{cm_mlperf_mil},
enable Loadgen support over the network for
MLPerf~\cite{cm_mlperf_network}, and create a universal, modular
Python-based harness to benchmark any AI/ML model using MLPerf Loadgen and
CM~\cite{cm_universal_python_harness}.

I also began collaborating with the Autonomous Vehicle Computing Consortium, 
where I created the CM4ABTF repository and helped prototype CM scripts to manage large Cognata datasets, 
train models, and run inference with the MLPerf Loadgen across a range of open-source 
and proprietary models, hardware, and software for OEMs and Tier 1 suppliers~\cite{cm4abtf}. 
This is an ongoing community effort, and everyone is welcome to join!

\section*{Auto-generated modular containers and READMEs with unified CM instructions}

I envision Collective Mind as a common, portable, and technology-agnostic
automation language that enables the co-design of more efficient and
cost-effective computer systems, similar to how high-level programming
languages translate and optimize programs into efficient,
hardware-specific code.

Furthermore, it will assist us in automatically generating optimized
containers, README files, and Jupyter notebooks with CM commands that
detail all the steps needed to build and run AI (virtualized MLOps and
DevOps). This will enable a modular and virtualized AI infrastructure that
can automatically adapt to any software and
hardware~\cite{cm_inference_readme_and_container}.

\section*{CM4Research: Collaboration with AI, ML and systems conferences}

Among the many reasons I am developing Collective Mind is to provide
a common language for automation, virtualization, and reproducibility that
can connect academia and industry, facilitating reproducible research and
technology transfer~\cite{acm_rep_23_cm_keynote}.

My goal is to provide a common interface that helps researchers,
engineers, and students encode all their knowledge and manage research
projects in a unified way, share and reuse automation recipes, reproduce
experiments, focus on innovation, and develop portable skills.

I have validated this approach with the\ community by prototyping a common
CM interface to manage several research projects published at ACM/IEEE
MICRO'23, NeurIPS, and other conferences and
journals~\cite{cm4research,ck-ae}. Additionally, I streamlined the process
of running the MLPerf inference benchmark during the Student Cluster
Competition at the ACM/IEEE SuperComputing conference~\cite{cm4scc23}
starting from 2022.

I am currently enjoying the opportunity to collaborate with academia and
industry to automate research projects using CM. Additionally, I am
helping students, researchers, and engineers learn how to modularize,
build, run, and optimize AI systems, including models, software, and
hardware.

\section*{Collective Knowledge Playground}

\begin{figure*}[]
  \centering
  \includegraphics[width=1.0\textwidth]{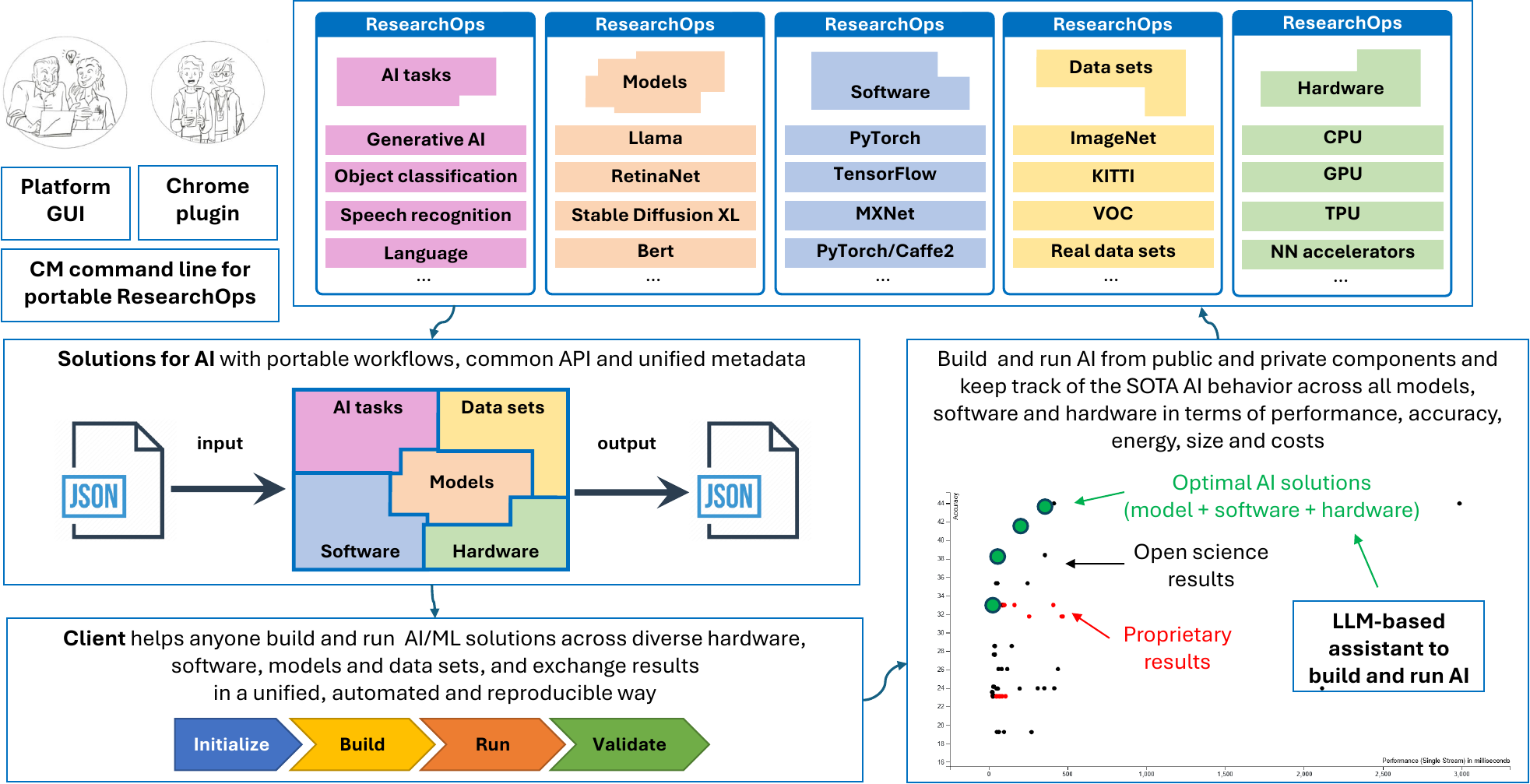}
  \caption{
   Collective Knowledge Playground: an educational community project to learn
   how to co-design more efficient and cost-effective software and hardware
   for AI, ML and other emerging workloads
   with the help of Collective Mind, virtualized MLOps, MLPerf 
   and reproducible optimization challenges.
  }
  \label{fig:ck-playground}
\end{figure*}

I am currently developing a platform called Collective Knowledge Playground (\href{https://access.cKnowledge.org}{access.cKnowledge.org}) 
to share and reuse CM scripts~\cite{cm_scripts} including automations to benchmark, optimize and co-design AI systems~\cite{cknowledge_howtorun}.
The platform aims to continuously test and stabilize these recipes across
diverse models, datasets, software, and hardware, while collecting
statistics about the behavior of AI systems. This is a collaborative
effort between academia and industry~\cite{cknowledge_experiments},
as conceptually illustrated in Figure~\ref{fig:ck-playground}.

My goal is to distribute the costly, lengthy, and repetitive process
of co-designing faster, cheaper, and more energy-efficient computer
systems for AI, ML, and other emerging workloads across many users.
By leveraging all the collected knowledge, we aim to model the behavior
of these systems and predict the most optimal configurations and
optimizations.

For example, we have collected various historical data from many MLPerf
benchmarks at the CK Playground in the CM
format~\cite{cknowledge_experiments}. This makes it easier
for everyone to visualize the data, add derived metrics such as energy
efficiency and costs~\cite{cknowledge_experiment_power_efficiency_example}, 
share optimization and reproducibility
reports~\cite{ck_playground_reproducibility_report_nvidia_orin}, and train
models to predict designs and suggest
CM commands~\cite{ck_playground_llm_assistant}.

Furthermore, one of my long-term goals for Collective Knowledge and Collective Mind is to help students 
co-design efficient software and hardware for AI and any other emerging workload 
in close collaboration with industry while solving the real world problems 
and acquiring portable skills, even as all components evolve continuously.

\section*{Future Plans}

We are just at the beginning of this community project, and there is still
much to be done. I continue improving Collective Mind,
Collective Knowledge Playground, MLPerf, and other related open-source
projects while using them to learn how to run AI, ML, and other popular
workloads in the most efficient and cost-effective way. My plans for
Collective Mind include adding cross-compilation, universal profiling,
full system optimization and co-design, enhanced support for ABTF,
MedPerf, MLPerf inference and training, TinyML, and MLPerf client.
Additionally, I aim to model AI systems' behavior, predict the most
profitable optimizations, and develop an LLM-based co-design assistant,
leveraging my previous related experience.

I am also developing the next generation of CM, called CMX (Common
Metadata eXchange), along with CMX4MLOps and other repositories. These
tools are designed to make it easier to develop and reuse new automations
across rapidly evolving open-source or proprietary models, datasets,
software, and hardware from different vendors and users. My ultimate goal
is to enable the automated and AI-based co-design of software and
hardware, auto-generated from the most suitable open-source and
proprietary components from various vendors. This will be based on user
requirements and constraints such as cost, latency, throughput, accuracy,
energy efficiency, size, and other critical factors.

I am inviting collaborators to join our Collective Mind developments 
at \href{https://github.com/mlcommons/ck}{github.com/mlcommons/ck}.

\section*{Acknowledgments}

I thank MLCommons and cTuning for their support of this community-driven project.

I am also deeply grateful to our contributors, colleagues and friends 
for their invaluable discussions, suggestions, contributions, and feedback
listed at {\it\href{https://github.com/mlcommons/ck/blob/master/CONTRIBUTING.md}{github.com/mlcommons/ck/blob/master/CONTRIBUTING.md}}
 
These include
@Henryfzh,
@Leonard226,
@Oseltamivir,
@Submandarine,
Resmi Arjun, 
Omar Benjelloun,
Alice Cheng,
Jiahao Chen,
Ramesh N Chukka,
Fabrice Costa,
Ray DeMoss,
Ryan T DeRue,
Himanshu Dutta,
Nicolas Essayan,
Justin Faust,
Diane Feddema,
Leonid Fursin,
Anirban Ghosh,
James Goel,
Michael Goin,
Jose Armando Hernandez,
Mehrdad Hessar,
Miro Hodak,
Sachin Idgunji,
Tom Jablin,
Nino Jacob,
David Kanter,
Alex Karargyris,
Georgios Karyotis,
Jason Knight,
Ilya Kozulin,
Steve Leak,
Amija Maji,
Peter Mattson,
Kasper Mecklenburg,
Pablo Gonzalez Mesa,
Thierry Moreau,
Sachin Mudaliyar,
Stanley Mwangi,
Ashwin Nanjappa,
Hai Ah Nam,
Nandeeka Nayak,
Datta Nimmaturi,
Lakshman Patel,
Arun Tejusve Raghunath Rajan,
Vijay Janapa Reddi,
Andrew Reusch,
Warren Schultz,
Sergey Serebryakov,
Byoungjun Seo,
Aditya Kumar Shaw,
Amrutha Sheleenderan,
Micah J Sheller,
Radoyeh Shojaei,
Ilya Slavutin,
Anandhu Sooraj,
Jinho Suh,
Arjun Suresh,
Badhri Narayanan Suresh,
David Tafur,
Chloe Tessier,
Gaurav Verma,
Zixian Wang,
Nathan Wasson,
Scott Wasson,
Haoyang Zhang,
Bojian Zheng
and Thomas Zhu.

I would also like to express my gratitude to Chloé Tessier for her
assistance with illustrations, as well as to Arjun Suresh and Anandhu
Sooraj for helping to maintain CM4MLOps repository, improve documentation 
and develop CM scripts to modularize MLPerf inference benchmarks
and automate MLPerf submissions. 

Lastly, I extend my gratitude to all the remarkable participants and reviewers who contributed 
to our optimization and reproducibility challenges:
{\it\href{https://access.cknowledge.org/playground/?action=contributors}{access.cKnowledge.org/playground/?action=contributors}}.

\bibliographystyle{plain}
\bibliography{cm}

\end{document}